\documentclass[10pt,twocolumn,letterpaper]{article}

\usepackage{iccv}
\usepackage{times}
\usepackage{epsfig}
\usepackage{graphicx}
\usepackage{amsmath}
\usepackage{amssymb}
\usepackage{algorithm}
\usepackage{algpseudocode}
\usepackage{amsmath,amssymb} 
\usepackage{xcolor}
\usepackage{multirow}
\usepackage{booktabs}
\usepackage{appendix}

\usepackage[pagebackref=true,breaklinks=true,letterpaper=true,colorlinks,bookmarks=false]{hyperref}

\iccvfinalcopy 

\def\iccvPaperID{1145} 

\ificcvfinal\fi

\begin{document}
	
	\title{Image Synthesis via Semantic Composition}
	
	\author{Yi Wang$^{1}$ \quad Lu Qi$^{1}$ \quad Ying-Cong Chen$^{2}$ \quad Xiangyu Zhang$^{3}$ \quad Jiaya Jia$^{1,4}$\\
		$^{1}$CUHK \quad$^{2}$HKUST \quad $^{3}$MEGVII Technology \quad $^{4}$SmartMore\\
		{\tt\small \{yiwang,luqi,leojia\}@cse.cuhk.edu.hk \ yingcong.ian.chen@gmail.com \ zhangxiangyu@megvii.com}
	}
	
	\maketitle
	\ificcvfinal\thispagestyle{empty}\fi

	\begin{abstract}
		In this paper, we present a novel approach to synthesize realistic images based on their semantic layouts.  It hypothesizes that for objects with similar appearance, they share similar representation. Our method establishes dependencies between regions according to their appearance correlation, yielding both spatially variant and associated representations. Conditioning on these features, we propose a dynamic weighted network constructed by spatially conditional computation (with both convolution and normalization). More than preserving semantic distinctions, the given dynamic network strengthens semantic relevance, benefiting global structure and detail synthesis.  We demonstrate that our method gives the compelling generation performance qualitatively and quantitatively with extensive experiments on benchmarks\footnote[1]{Project page is at \href{https://shepnerd.github.io/scg/}{https://shepnerd.github.io/scg/}}. 
	\end{abstract}
	
	\section{Introduction}
\begin{figure}[t]
		\begin{center}
			\centering
			\includegraphics[width=0.95\linewidth]{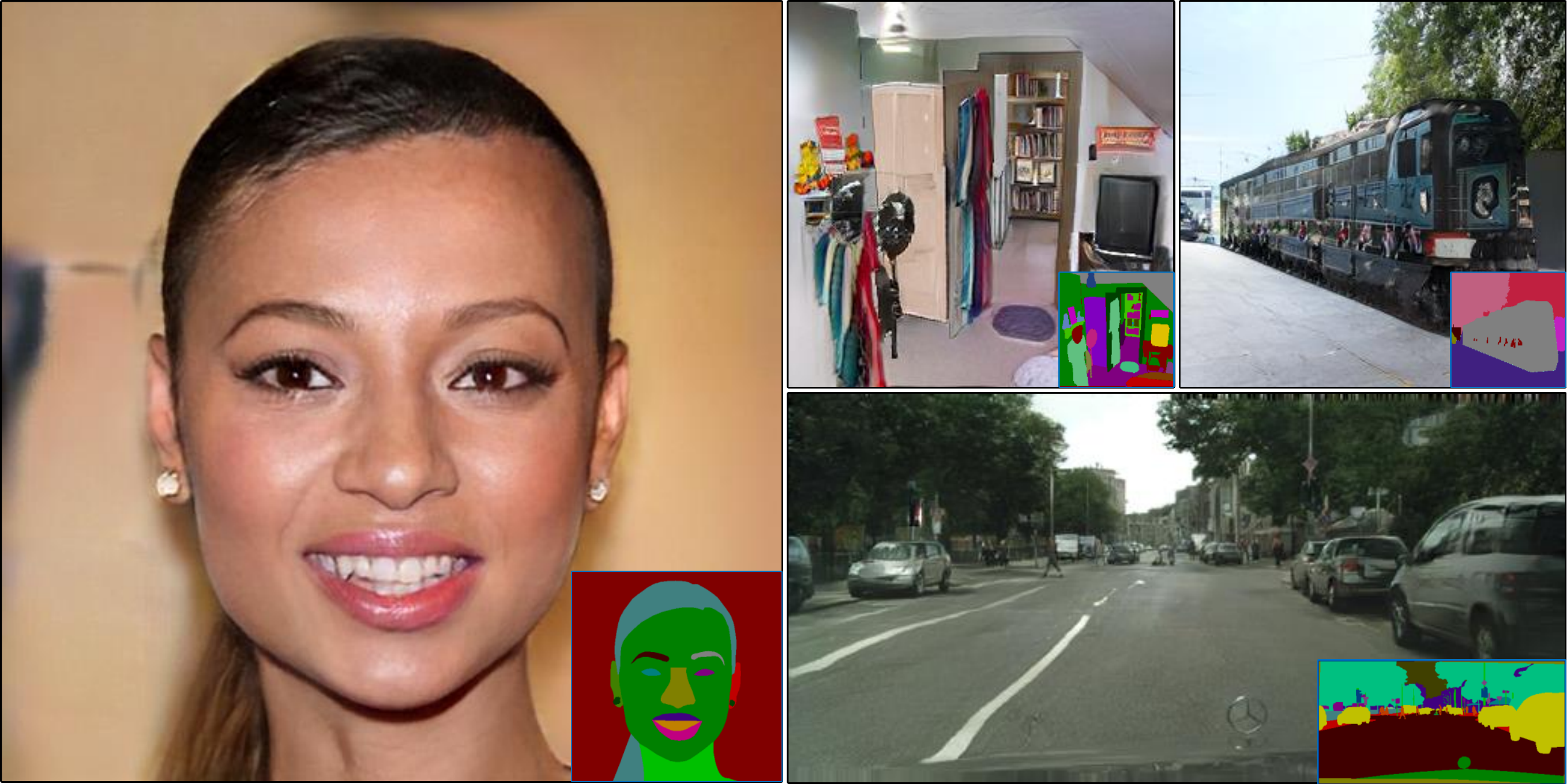} 
		\end{center}
		\caption{Semantic image synthesis results of our method on face and scene datasets.} 
		\label{fig_teaser}
		\vspace{-0.1in}
	\end{figure}

	Semantic image synthesis transforms the abstract semantic layout to realistic images, an inverse task of semantic segmentation (Figure \ref{fig_teaser}). It is widely used in image manipulations and content creation. Recent methods on this task are built upon generative adversarial networks (GAN) \cite{goodfellow2014generative}, modeling image distribution conditioning on segmentation masks.  
	
	Despite its substantial achievement \cite{isola2017image,wang2018high,park2019semantic,liu2019learning,tang2019multi,chen2017photographic,qi2018semi,jiang2020tsit,tang2020local}, this line of research is still challenging due to the high complexity of characterizing object distributions. Recent advance \cite{park2019semantic,liu2019learning} on GAN-based image synthesis concentrates on how to exploit spatial semantic variance in the input for better preserving layout and independent semantics, leading to further generative performance improvement. They both use different parametric operators to handle different objects. 
	
	Specifically, SPADE \cite{park2019semantic} proposes a spatial semantics-dependent denormalization operation for the common normalization, as the feature statistics are highly correlated with semantics. CC-FPSE \cite{liu2019learning} extends this idea to convolution using dynamic weights, generating spatially-variant convolutions from the semantic layouts. Relationships between objects are implicitly modeled by the weight-sharing convolutional kernels (SPADE) or hierarchical structures brought by stacked convolutions. We believe when performing semantic-aware operations, enhancing object relationship could be further beneficial to final synthesis, since context and long-range dependency have proven effective in several vision tasks \cite{dai2017deformable,wang2018non,zhang2018self,yu2018generative,wang2020attentive}.
	
	To explicitly build connection between objects and stuff in image synthesis, we propose a semantic encoding and stylization method, named semantic-composition generative adversarial network (SC-GAN). We first generate the \textit{semantic-aware} and \textit{appearance-correlated} representations from mapping the discrete semantic layout to their corresponding images. 
	
	Our idea is inspired by the following observation. Different semantics are with labels in scene or face parsing datasets. Some of them are highly correlated in appearance, e.g., the left and right eyes in CelebAMask-HQ \cite{lee2020maskgan}. We abstract objects in images or feature maps into vectors by encoding the semantic layout, which we call semantic vectors. This intermediate representation is to characterize the relationship between different semantics based on their appearance similarity.
	
	With these semantic vectors, we create \textit{semantic-aware and appearance-consistent local operations in a semantic-stylization fashion}. Consistent with the proposed semantic vectors, these dynamic operators are also variant in semantics and correlated in appearance. Our proposed operators are extended from existing conditional computation \cite{yang2019condconv} to stylize the input conditioned on the semantic vectors. Specifically, we exploit semantic vectors to combine a shared group of learnable weights, parameterizing the convolutions and normalizations used in semantic stylization. 
	
	Note the learning of semantic vectors and render candidates is non-trivial. Intuitive designs that encode the semantic layout to semantic vectors for later dynamic computations make semantic vectors stationary, since these semantic vectors are not directly regularized by the appearance information in the image. In this paper, we make the learning of semantic encoding and stylization relatively independent. Semantic encoding is trained by estimating the corresponding natural images from the input semantic layouts by maximum likelihood. Semantic stylization is trained in an adversarial manner. 
	
	Our method is validated on several image synthesis benchmark datasets quantitatively and qualitatively. Also, the decoupled design of semantics encoding and stylization makes our method applicable to other tasks, e.g.,  unpaired image translation \cite{zhu2017unpaired}.
	Our contribution is threefold.
	\begin{itemize}
		\vspace{-0.05in}
		\item We propose a new generator design for GAN-based semantic image synthesis. We present spatially-variant and appearance-correlated operations (both convolution and normalization) via spatially conditional computation, exploiting both local and global characteristics for semantic-aware processing.
		\vspace{-0.05in}
		\item Our proposed generator with a compact capacity outperforms other popular ones on face and scene synthesis datasets in visual and numerical comparisons. 
		\vspace{-0.05in}
		\item The decoupled design in our method finds other applications, \eg\, unpaired image-to-image translation.
	\end{itemize}
	
	\begin{figure*}[t]
		\begin{center}
			\centering
			\includegraphics[width=0.9\linewidth]{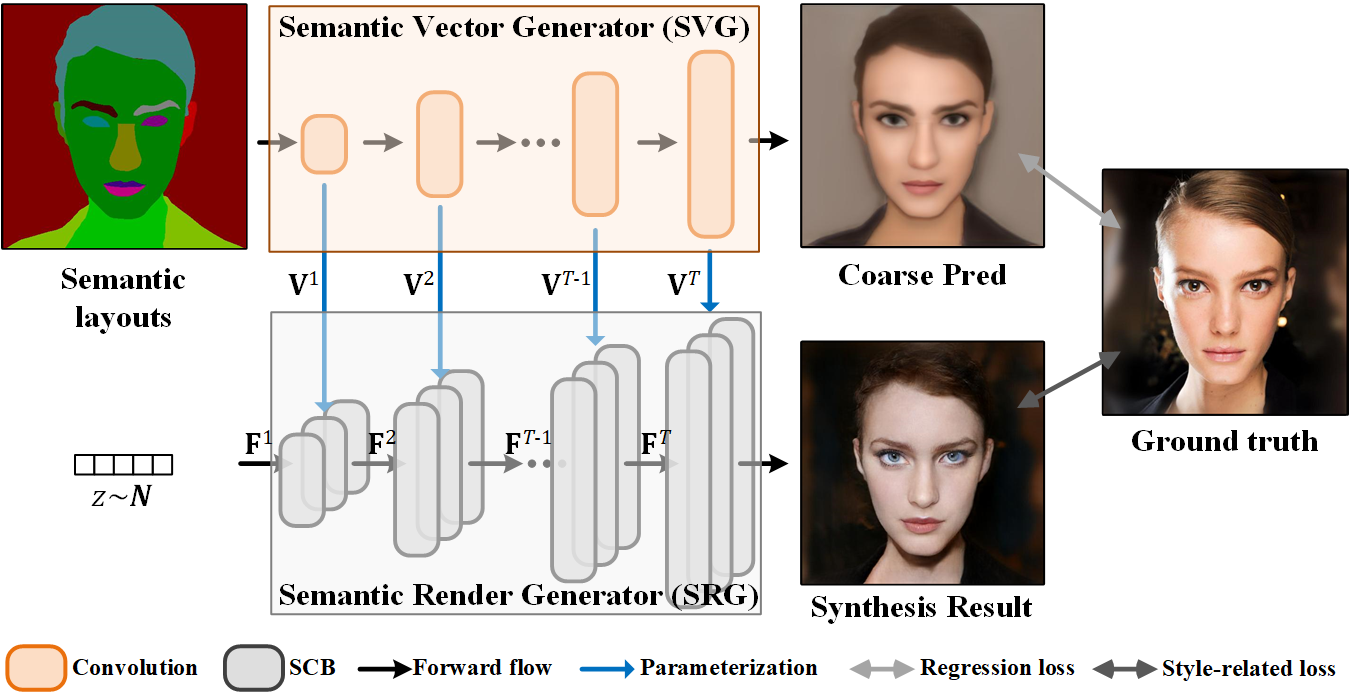} 
		\end{center}
		\vspace{-0.1in}
		\caption{Framework of SC-GAN. SCB denotes the spatially conditional block. It is constructed by spatial conditional convolutions and normalizations, parameterized by semantic vectors $\mathbf{V}$. Its design is given in Figure \ref{fig_residual} and Section \ref{sec_srg}.}
		\label{fig_frame}
		\vspace{-0.1in}
	\end{figure*}
	
	\section{Related Work}
	
	\subsection{Semantic Image Synthesis}
	This task is to create a realistic image based on the given semantic layout (pixel-level semantic labels). Essentially, it is an ill-posed problem as one semantic layout may correspond to several feasible pictures. It can be dated back to `Image analogy' in 2001 \cite{hertzmann2001image}, in which the mentioned mapping uncertainties are resolved by the local matching and constraints from a reference image. 
	
	Recent learning-based approaches \cite{isola2017image,wang2018high,wang2018inpainting,park2019semantic,liu2019learning,tang2019multi,wang2019wide,chen2017photographic,qi2018semi,jiang2020tsit,tang2020local,wang2020vcnet,zhu2020sean,sushko2020you} greatly advance this area, formulating it as an image distribution learning problem conditioned on the given semantic maps. Due to the development of conditional generative adversarial networks (cGAN) \cite{mirza2014conditional}, pix2pix makes seminal exploration on image synthesis \cite{isola2017image}. It gives some components and principles about how to apply cGAN to this problem, including loss design, generator structures (e.g., encoder-decoder and U-Net), and Markovian discriminator (also known as PatchGAN). 
	
	Later, Wang \etal\ propose a new enhanced version pix2pixHD \cite{wang2018high}. By introducing a U-Net style generator with a larger capacity and several practical techniques for improving GAN training, including feature matching loss, multi-scale discriminators, and perceptual loss, their method boosts the image synthesis performance on producing vivid details. Further, SPADE is developed on improving the realism of the synthesized images by working on the normalization issue \cite{park2019semantic}. It shows using the given segmentation masks to explicitly control the affine transformation in the used normalization can better preserve the generated semantics, leading to a noticeable improvement. Such an idea is further extended in CC-FPSE \cite{liu2019learning}. It dynamically generates convolutional kernels in the generator conditioning on the input semantics. 
	
	Besides of GAN-based methods, research explores this task from other perspectives. Chen \etal\ \cite{chen2017photographic} produce images using cascaded refinement networks (CRN) progressively with regression, starting from small-scale semantic layouts. Qi \etal\ \cite{qi2018semi} proposed a semi-parametric approach to directly utilize object segments in the training data. They retrieve object segments with the same semantics and similar shapes from the training set to fill the given semantic layout, and then regress these assembled results to the final images. Additionally, Li \etal\ \cite{li2019diverse} employ implicit maximum likelihood estimation to CRN, to pursue more diverse results from a semantic layout.
	
	\subsection{Dynamic Computation}
	In the development of neural network components, dynamic filters \cite{jia2016dynamic} or hypernetworks \cite{ha2016hypernetworks} are proposed for their flexibility to input samples. It generates dynamic weights conditioned on the input or input-related features for parameterizing some operators (mostly fully-connected layers or convolutions). This has been applied to many tasks \cite{yang2018metaanchor,qi2020pointins,hu2019meta,chen2020dynamic,park2019semantic,liu2019learning,jo2018cvpr}.
	
	\vspace{-0.15in}
	\paragraph{Conditionally Parameterized Convolutions} It is a special case of dynamic filters, which produces dynamic weights by combining the provided candidates conditionally \cite{yang2019condconv}. It uses the input features $\mathbf{X}$ to generate the input-dependent convolution kernels in the neural nets by a learnable linear combination of a group of kernels $\{\mathbf{w}\}_{1,...,n}$, as $(\sum_{i=1}^{n}\alpha_i(\mathbf{X}) \mathbf{w}_i)$, where $\alpha_i(\mathbf{X})$ computes an input-dependent scalar to choose the given kernel candidates, and $n$ is the expert number. It is equivalent to a linear mixture of experts, with more efficient implementation.
	
	In this paper, our proposed dynamic computation units are extended from conditionally parameterized convolutions. We generalize the scalar condition into a spatial one and also apply these techniques to normalization.
	
	\section{Semantic-Composition GAN}
	
	We aim to transform a semantic layout $\mathbf{S} \in \{0, 1\}^{h \times w \times c}$ to a realistic picture $\mathbf{\hat{I}} \in \mathbb{R}^{h \times w \times 3}$ (where $h$, $w$, and $c$ denote the height, width, and the category number in semantic layout, respectively), as $\mathbf{\hat{I}} = f (\mathbf{S})$, in which $f$ is a nonlinear mapping. During training, paired-wise data is available given the corresponding natural image $\mathbf{I}$ of $\mathbf{S}$ provided. We demand synthesized image $\mathbf{\hat{I}}$ to match the given semantic layout. But $\mathbf{\hat{I}}$ and $\mathbf{I}$ are not necessarily identical. 
	
	Our semantic-composition GAN (SC-GAN) decouples semantic image synthesis into two parts: semantic encoding and stylization. They are realized by semantic vector generator $G_{V}$ (SVG) and semantic render generator $G_{R}$ (SRG), respectively. As shown in Figure \ref{fig_frame}, SVG takes the semantic layout $\mathbf{S}$ and produces multi-scale semantic vectors in a feature map form (since we treat each feature point as a semantic vector). SRG is to transform a random sampled noise to the final synthesized image with a dynamic network. The key operators (convolution and normalization) in this network are conditionally parameterized by the semantic vectors provided by SVG and a group of weight candidates.
	
	\subsection{Semantic Vectors Generation} \label{sec_svg}
	SVG is to transform the input discrete semantic labels $\mathbf{S}$ into semantic-and-appearance correlated representation of semantic vectors, building the relationship between different semantics according to their appearance similarities. For example, grass and trees are represented by different semantic labels in COCO-stuff \cite{caesar2018coco}, sharing similarities in color and texture. Our method represents their corresponding regions with different but similar representations.
	
	Generally, semantic vectors are learned from encoding of the input semantic labels into the corresponding image. SVG takes input of the semantic layout $\mathbf{S}$ and generates the corresponding semantic vectors in feature map form as $\mathbf{V} \in [0, 1]^{h' \times w' \times n}$  with different scales. It is expressed as
	\begin{equation}
		\{\mathbf{V}^t\}_{t=1,...,T} = f_{\text{V}}(\mathbf{S}),
	\end{equation}
	where $t$ denotes a different spatial scale. 
	
	SVG is in a cascaded refinement form, structured as CRN \cite{chen2017photographic}. We feed a small-scale semantic layout into it, encode the input and upsample the features, and concatenate it with a larger-scale semantic layout. We repeat this process until the output reaches the final resolution. 
	
	\vspace{-0.15in}
	\paragraph{Nonlinear Mapping} We regularize the computed semantic vectors using a nonlinear mapping. Directly employing the these unconstrained vectors would lead to performance drop (shown in Section \ref{sec_ab}). Suppose $\mathbf{v}=\mathbf{V}_{i,j} \in \mathbb{R}^{n}$, where $i$ and $j$ index height and width. We further normalize its values into $[0, 1]$ with softmax for better performance and interpretability as
	\begin{equation} \label{eq_softmax}
		g(\tau \mathbf{v})_i = \frac{\exp(\tau \mathbf{v}_i)}{\sum_{j=1}^{n}\exp(\tau \mathbf{v}_j)},
	\end{equation}
	where $\mathbf{v}_i$ denotes the $i$th scalar in $\mathbf{v}$, and $\tau=0.05$ is the temperature to control smoothness of the semantic vectors $\mathbf{v}$. The smaller $\tau$ is, the smoother $\mathbf{v}$ is. Note $g(\cdot)$ could be sigmoid, tanh, or other nonlinear functions. The performance along with our choices will be empirically compared and analyzed in the Section \ref{sec_ab}.
	
	\begin{figure}[t]
		\begin{center}
			\centering
			\setlength{\tabcolsep}{1mm}{
				\begin{tabular}{cc}
					\includegraphics[width=0.46\linewidth]{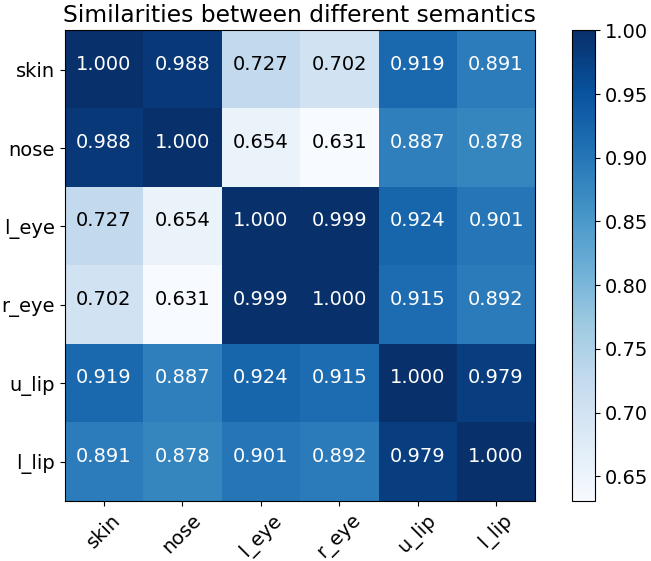} &
					\includegraphics[width=0.50\linewidth]{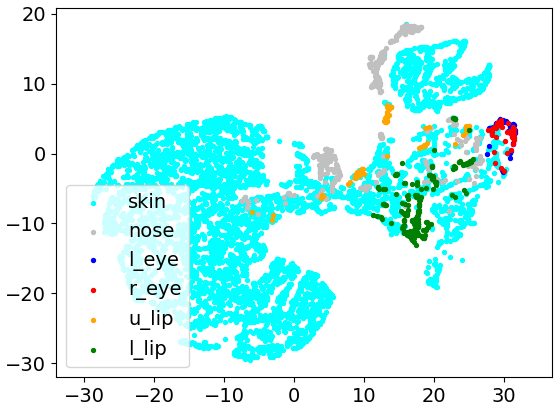} \\
				\end{tabular}
			}
		\end{center}
		\vspace{-0.1in}
		\caption{Feature correlation matrix between semantic vectors of different semantics (left) and 2D feature distributions (compressed by t-SNE) of semantic vectors (right) on CelebAMask-HQ.}
		\label{fig_tsne_face}
	\end{figure}
	
	\vspace{-0.15in}
	\paragraph{Effectiveness of Semantic Vectors} 	We visualize the correlation between different semantic regions using semantic vectors. It is found that these vectors are related by the appearance similarity. Figure \ref{fig_tsne_face} (left) shows cosine similarities between mean semantic vectors of different semantics. Note the semantic vectors representing the left and right eyes are almost identical (with cosine similarity 0.999). This is also observed in the relationship between the upper and lower lips. Intriguingly, it also reveals that the semantic vectors for skin are close to those of the nose. 
	
	Figure \ref{fig_tsne_face} (right) illustrates how these semantic vectors are distributed (compressed to 2D by t-SNE, only 1\% points are visualized for clarity). Note that blue points standing for left eyes are much overlapped with those of right eyes. Also, the orange point cluster is close to the green one for upper and lower lips. It validates that semantic vectors can give similar representations to different semantic regions with a similar appearance. These semantic vectors are extracted from 100 random images of CelebAMask-HQ.  
	
	\begin{figure}[t]
		\begin{center}
			\centering
			\includegraphics[width=0.9\linewidth]{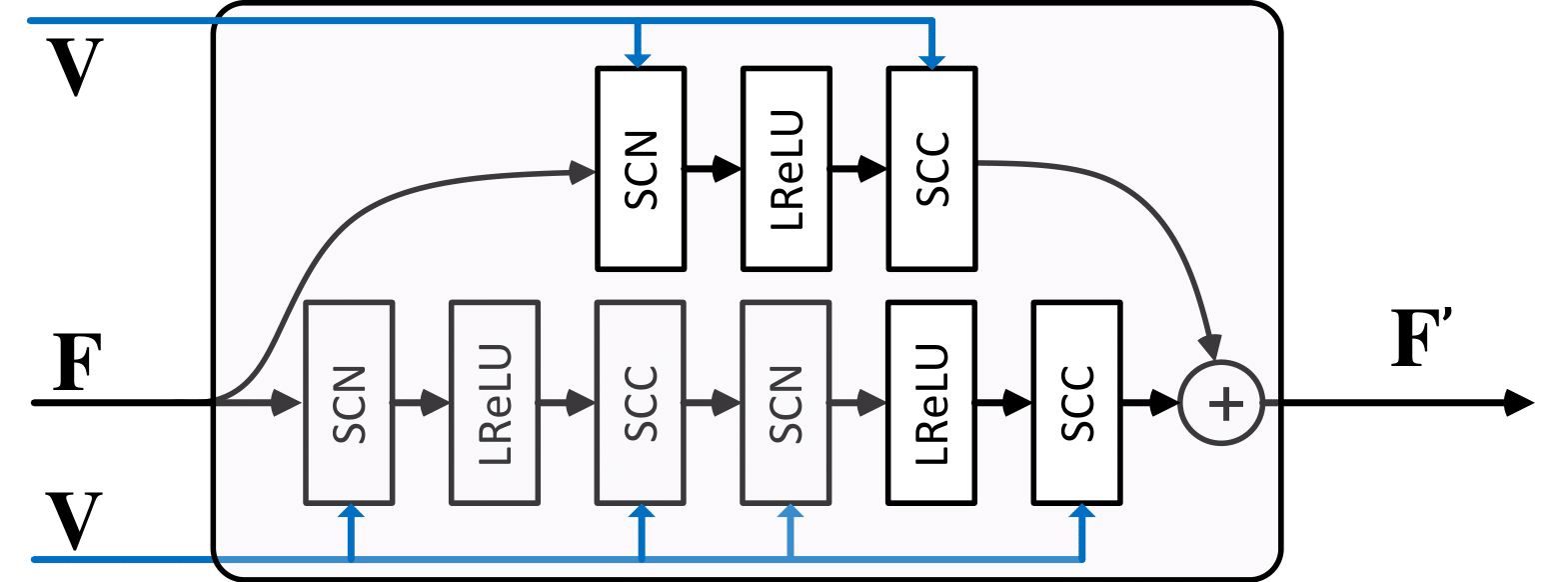} 
		\end{center}
		\vspace{-0.1in}
		\caption{Employed residual block using the proposed spatially conditional convolution (SCC) and normalization (SCN).}
		\label{fig_residual}
		\vspace{-0.1in}
	\end{figure}
	
	\subsection{Semantic Render Generation} \label{sec_srg}
\begin{figure}[t]
		\begin{center}
			\centering
			\begin{tabular}{cc}
				\includegraphics[width=0.4\linewidth]{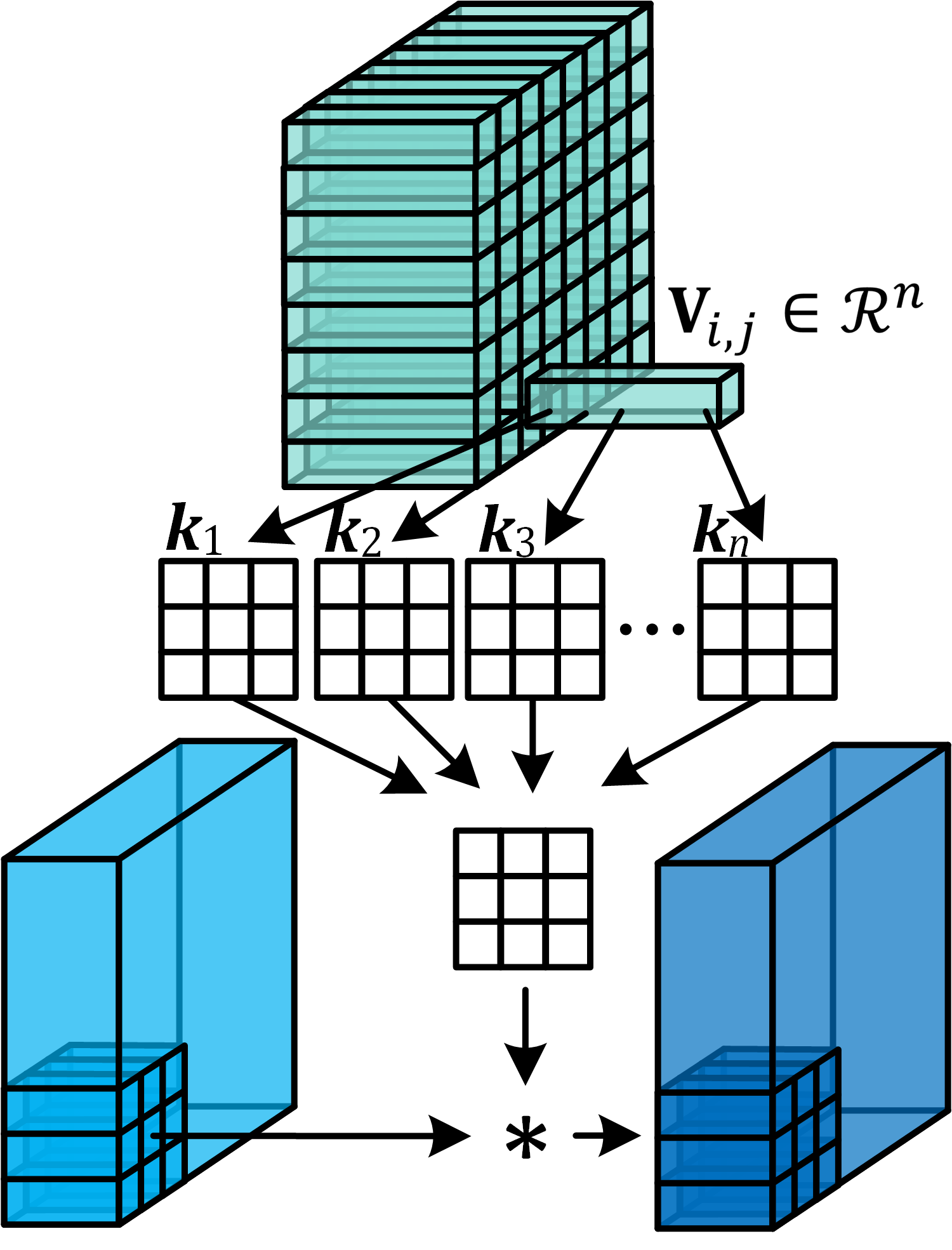} &
				\includegraphics[width=0.56\linewidth]{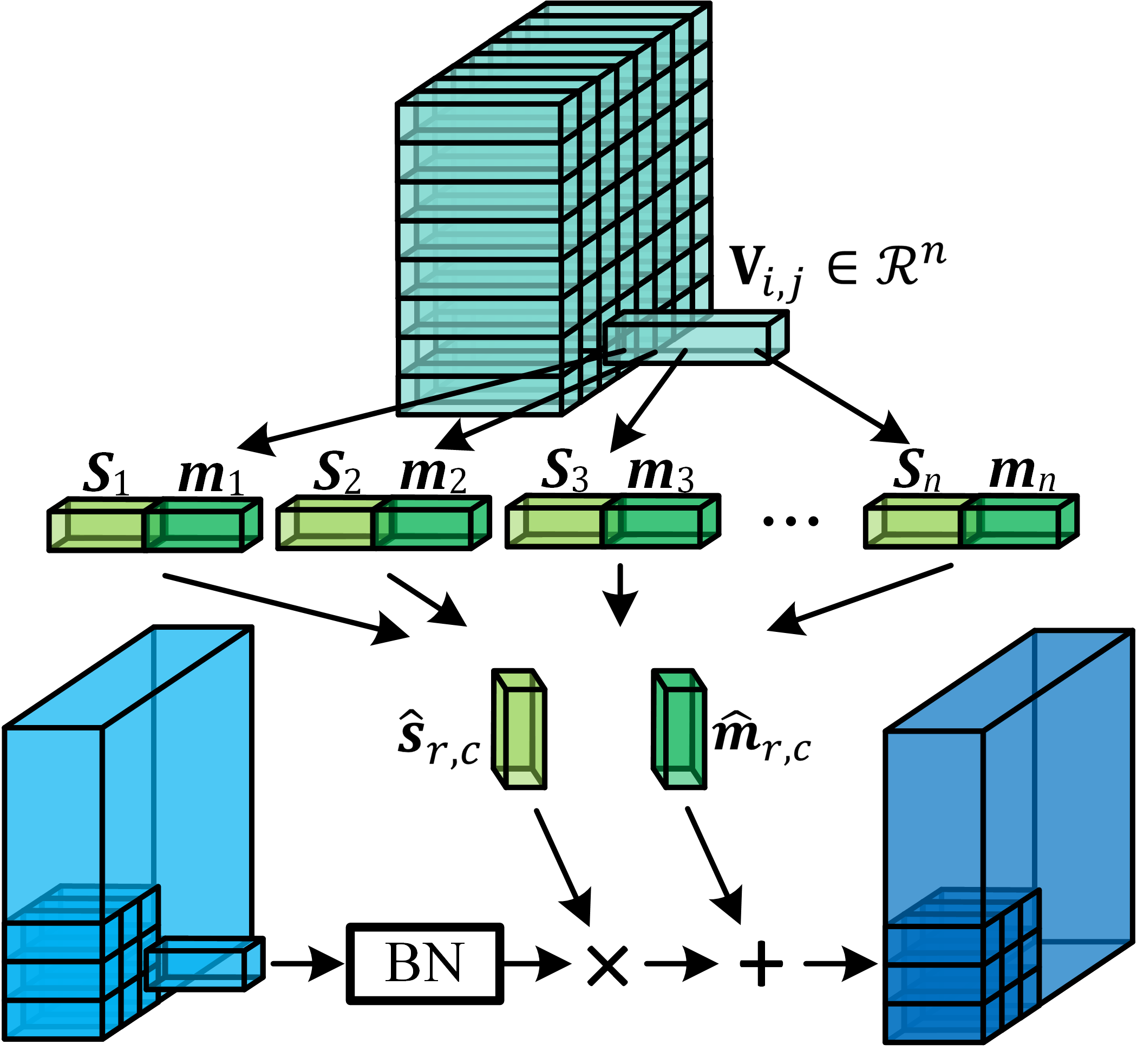} \\
			\end{tabular}
		\end{center}
		\vspace{-0.1in}
		\caption{Conceptual illustration of spatial conditional computation. Left: spatially conditional convolution (SCC), right: spatially conditional normalization (SCN).}
		\label{fig_scc}
		\vspace{-0.1in}
	\end{figure}

	SRG is a generator in a progressive manner, built with the residual block \cite{he2016deep} (Figure \ref{fig_residual}) formed by our proposed spatial conditional convolution (SCC, left of Figure \ref{fig_scc}) and spatial conditional normalization (SCN, right of Figure \ref{fig_scc}). It transforms random noise $z$ into the target image $\mathbf{\hat{I}}$, conditioned on $\{\mathbf{V}^t\}$ of $\mathbf{S}$. It is formulated as
	\begin{equation}
		\mathbf{\hat{I}} = f_{\text{R}} (z | \{\mathbf{V}^t\}), 
	\end{equation}
	where $z \sim \mathcal{N}$ and $\mathcal{N}$ is a standard normal distribution.
	
	Both SCC and SCN are spatially conditional, parameterized from semantic vectors and a shared group of weights, making them produce semantic-aware and appearance-correlated operators. Their designs are detailed below.
	
	\vspace{-0.15in}
	\paragraph{Spatially Conditional Convolution}
	For input feature maps $\mathbf{F} \in \mathbb{R}^{h \times w \times c}$ (intermediate representation of $z$ in SRG), assuming its learnable kernel candidates are $\{\mathbf{k}_1, \mathbf{k}_2, ..., \mathbf{k}_n\}$, with semantic vectors $\mathbf{V} \in \mathbb{R}^{h \times w \times n}$, we compute the corresponding output as
	\begin{equation} \small \label{eq_scc}
		\text{scc}(\mathbf{F}, \mathbf{V}; \{\mathbf{k}_i\}_{i=1,...,n})_{r,c} = \sum_{i=1}^{n}(\mathbf{V}_{r,c,i}\mathbf{k}_i) * \mathbf{F}_{r,c},
	\end{equation}
	where $r$ and $c$ indicate the row and column indexes of the given feature maps, and $i$ indexes both the channel of $\mathbf{V}$ and the weight candidate.
	
	\vspace{-0.15in}
	\paragraph{Spatially Conditional Normalization}
	Similarly, the spatially conditional normalization employs linearly combined mean and variance for the affine transformation after normalization. Still for $\mathbf{F}$ and $\mathbf{V}$, suppose its learnable mean and variance candidates are $\{\mathbf{m}_1, \mathbf{m}_2, ..., \mathbf{m}_n\}$ and $\{\mathbf{s}_1, \mathbf{s}_2, ..., \mathbf{s}_n\}$, we yield the normalized output as
	\begin{equation} \small \label{eq_scn}
		\begin{split}
			\text{scn}(& \mathbf{F}, \mathbf{V}; \{\mathbf{m}_i, \mathbf{s}_i\})_{r,c} = \frac{\mathbf{F}_{r,c}-\mu(\mathbf{F})}{\sigma(\mathbf{F})} \times \mathbf{\hat{s}}_{r,c} + \mathbf{\hat{m}}_{r,c},\\
			\text{where} & \quad \mathbf{\hat{s}}_{r,c} =\sum_{i=1}^{n}(\mathbf{V}_{r,c,i}\mathbf{s}_i), \;
			\mathbf{\hat{m}}_{r,c}=\sum_{i=1}^{n}(\mathbf{V}_{r,c,i}\mathbf{m}_i),
		\end{split}
	\end{equation}
	where $\mu(\cdot)$ and $\sigma(\cdot)$ are to compute the mean and standard variance of their input, respectively. Mean and variance candidates are initialized to 0 and 1, respectively.
	
	\vspace{-0.15in}
	\paragraph{Analysis}
	Inheriting from the spatially adaptive processing idea from SPADE \cite{park2019semantic} in semantic image synthesis, our model generates semantic-aware convolutions and normalization to handle different semantic regions indicated by the input. This idea is also explored in CC-FPSE \cite{liu2019learning}, which employs segmentation masks to parameterize the conditional convolution directly by generating weights. Although it improves the generation quality of SPADE, the computation is expensive, since its independent spatially variant convolutional operations can only be implemented using local linear projection instead of standard convolution. 
	
	In contrast, our operators are correlated when they cope with regions with similar appearance and yet with different labels, achieved by the semantic vectors and the shared group of weights. Besides, our SCC utilizes standard convolutions for efficient training and evaluation.
	As validated in our experiments, our design is beneficial to long-range dependency modeling, consistently improving generation performance.
	
	\subsection{Learning Objective} \label{sec_obj}
	The optimization goal of our method consists of two parts with style-related loss and regression loss. The former contains perceptual loss, GAN loss, and feature matching loss. Regression loss on SVG is to learn semantic-aware and appearance-consistent vectors, named semantic vector generation loss. In general, our optimization target is
	\begin{equation}
		\mathcal{L} = \lambda_{p}\mathcal{L}_{p} + \lambda_{gan} \mathcal{L}^{G}_{gan} + \lambda_{fm} \mathcal{L}_{fm} + \lambda_{s} \mathcal{L}_{s},
	\end{equation}
	where $\lambda_{p}$, $\lambda_{gan}$, $\lambda_{fm}$, and $\lambda_{rwg}$ are trade-off regularization parameters, set to 10, 1, 10, and 2, respectively.
	
	\vspace{-0.15in}
	\paragraph{Perceptual Loss}
	We employ the pretrained natural image manifold to constrain the generative space of our model. Specifically, we minimize the discrepancy between the produced images and their corresponding ground truth in the feature space of VGG19 \cite{simonyan2014very} as
	\begin{equation} \label{eq_perce}
		\mathcal{L}_{p}(\mathbf{\hat{I}}, \mathbf{I}) = \sum^{5}_{i=1}||\Phi_{i\_1}(\mathbf{\hat{I}}) - \Phi_{i\_1}(\mathbf{I})||_1,
	\end{equation}
	where $\Phi_{i\_1}$ indicates extracting the feature maps from the layer ReLU $i\_1$ of VGG19. 
	
	\vspace{-0.15in}
	\paragraph{GAN Loss}
	We train our generator and discriminator using hinge loss. Its learning goal on generator is
	\begin{equation}
		\mathcal{L}_{gan}^{G} =  -\text{E}_{z \sim \mathbb{P}_z, \mathbf{S} \sim \mathbb{P}_\text{data}}D(G(\mathbf{S}|z)),
	\end{equation}
	where $G$ and $D$ denote the generator (including SVG and SRG) and discriminator of our SC-GAN, respectively. The corresponding optimization goal for the discriminator is
	\begin{equation}
		\begin{split}
			\mathcal{L}_{gan}^{D} = &\text{E}_{\mathbf{I} \sim \mathbb{P}_\text{data}}[\text{max}(0, 1-D(\mathbf{I}))]+\\ &\text{E}_{z \sim \mathbb{P}_z, \mathbf{S} \sim \mathbb{P}_\text{data}}[\text{max}(0, 1+D(G(\mathbf{S}|z))].
		\end{split}
	\end{equation}
	For the discriminator, we directly employ the feature pyramid semantics-embedding discriminator from \cite{liu2019learning}. 
	
	\vspace{-0.15in}
	\paragraph{Semantic Vector Generation Loss} To prevent SVG generates trivial semantic vectors for later generation, we regularize its learning by predicting the corresponding information of the fed semantic layout as
	\begin{equation} \label{eq_svg}
		\mathcal{L}_{s} = ||f_\text{V}^\text{out}(\mathbf{S})-\mathbf{I}||_p,
	\end{equation}
	where $f_\text{V}^\text{out}(\mathbf{S})$ denotes the predicted image from SVG. $p=1$ or $2$. Note that we can also conduct such measure in the perceptual space of a pretrained classification network like Eq. (\ref{eq_perce}). This is studied in Section \ref{sec_ab}.
	
	\begin{table*}[h]
		\centering
		\caption{Quantitative results on the validation sets from different methods.}
		\label{tb_me_evaluation1}
		\setlength{\tabcolsep}{1mm}{
			\begin{tabular}{c|ccccccccccc}
				\toprule
				\multirow{2}*{Method} & \multicolumn{2}{c}{CelebAMask-HQ} & \multicolumn{3}{c}{Cityscapes} & \multicolumn{3}{c}{ADE20K} & \multicolumn{3}{c}{COCO-Stuff}  \\
				~ & FID $\downarrow$ & LPIPS $\downarrow$ & mIoU $\uparrow$ & Acc $\uparrow$ & FID $\downarrow$ & mIoU $\uparrow$ & Acc $\uparrow$ & FID $\downarrow$ & mIoU $\uparrow$ & Acc $\uparrow$ & FID $\downarrow$ \\
				\midrule
				CRN \cite{chen2017photographic}& N/A & N/A & 52.4 & 77.1 & 104.7 & 22.4 & 68.8 & 73.3  & 23.7 & 40.4 & 70.4 \\
				SIMS \cite{qi2018semi}& N/A & N/A  & 47.2 & 75.5 & 49.7 & N/A & N/A & N/A  & N/A & N/A & N/A \\
				pix2pixHD \cite{wang2018high} & 54.7 & 0.529 & 58.3 & 81.4 & 95.0 & 20.3 & 69.2 & 81.8 & 14.6 & 45.7 & 111.5 \\
				SPADE \cite{park2019semantic} & 42.2 & 0.487  & 62.3 & 81.9 & 71.8 & 38.5 & 79.9 & 33.9 & 37.4 & 67.9 & 22.6\\
				CC-FPSE \cite{liu2019learning} & N/A & N/A  & 65.6 & 82.3 & 54.3 & 43.7 & 82.9 & 31.7 & 41.6 & 70.7 & 19.2\\
				Ours & \textbf{19.2} & \textbf{0.395}  & \textbf{66.9} & \textbf{82.5} & \textbf{49.5} & \textbf{45.2} & \textbf{83.8} & \textbf{29.3} & \textbf{42.0} & \textbf{72.0} & \textbf{18.1} \\
				\bottomrule
			\end{tabular}
		}
		\vspace{-0.1in}
	\end{table*}
	
	\subsection{Implementation}
	We apply spectral normalization (SN) \cite{miyato2018spectral} both on the generator and discriminator. Also, a two time-scale update rule \cite{heusel2017gans} is used during training. The learning rates for the generator and discriminator are $1e-4$ and $4e-4$, respectively. The training is conducted with Adam \cite{kingma2014adam} optimizer with $\beta_1=0$ and $\beta_2=0.999$. For the used batch normalization, all statistics are synchronized across GPUs. Unless otherwise specified, the used candidate number $n$ from Eqs. \eqref{eq_scc} and \eqref{eq_scn} of our method is set to 3 in experiments.
	
	\section{Experiments}
	
	Our experiments are conducted on four face and scene parsing datasets: CelebAMask-HQ \cite{lee2020maskgan}, Cityscapes \cite{cordts2016cityscapes}, ADE20K \cite{zhou2017scene}, and COCO-Stuff \cite{caesar2018coco}. In our experiments, images and their corresponding semantic layouts in CelebAMask-HQ, Cityscapes, ADE20K, and COCO-Stuff are resized and cropped into $512 \times 512$, $256 \times 512$, $256 \times 256$, $256 \times 256$, respectively. The train/val splits follow the setting of these datasets.
	
	For training of our method, we take 100, 200, 200, and 100 epochs on CelebAMask-HQ, Cityscapes, ADE20K, and COCO-Stuff, respectively. The first half of epochs on the first three datasets are with full learning rates and the remaining half linearly decays with learning rates approaching 0. Our computational platform is with 8 NVIDIA 2080Ti GPUs.
	
	\vspace{-0.15in}
	\paragraph{Baselines}  We take CRN \cite{chen2017photographic}, SIMS \cite{qi2018semi}, Pix2pixHD \cite{wang2018high}, SPADE \cite{park2019semantic}, Mask-GAN \cite{lee2020maskgan}, and CC-FPSE \cite{liu2019learning} as baselines. The following evaluation is all with their original implementation and pretrained models from their official release. Some new results on CelebAMask-HQ from SPADE are trained from scratch with their default training setting. CC-FPSE is not applicable here since it consumes more GPU memory on the face dataset than we can afford. The reason is that it realizes conditional convolution with spatially independent local linear processing, leading to considerable computational overhead. Note that Mask-GAN is developed for facial image manipulation in style transfer, which exploits both the semantic layout and a reference image. Its qualitative and quantitative evaluation is only for reference because it uses extra input information.
	
	
	About the model complexity, our SC-GAN has the minimal capacity of 66.2M parameters, compared to referred numbers of parameters of 183.4M for Pix2pixHD, 93.0M for SPADE, and 138.6M for CC-FPSE.
	
	\vspace{-0.15in}
	\paragraph{Evaluation Metrics} Following the testing protocol in the semantic image synthesis task \cite{wang2018high,park2019semantic}, we evaluate our methods along with baselines in the following perspectives: quantitative performance in Fr{\'e}chet Inception Score (FID) \cite{heusel2017gans} and learned perceptual image patch similarity (LPIPS) \cite{zhang2018unreasonable}, semantic segmentation, and user study. 
	In semantic segmentation, like the existing work \cite{wang2018high,park2019semantic}, we use mIoU and mAcc (mean pixel accuracy) performed on the synthesized results from the trained segmentation models to assess the result quality. DeepLabV2 \cite{chen2017deeplab}, UperUnet101 \cite{xiao2018unified}, and DRN-D-105 \cite{yu2017dilated} are employed for COCO-Stuff, ADE20K, and Cityscapes, respectively.
	
	\begin{figure*}[!ht]
		\begin{center}
			\includegraphics[width=0.95\linewidth]{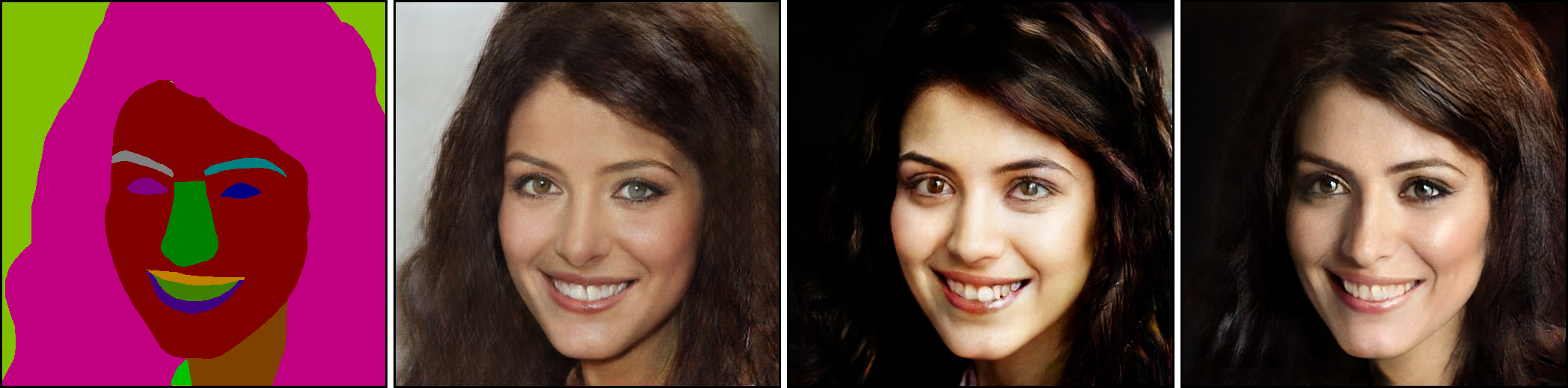} 
			\includegraphics[width=0.95\linewidth]{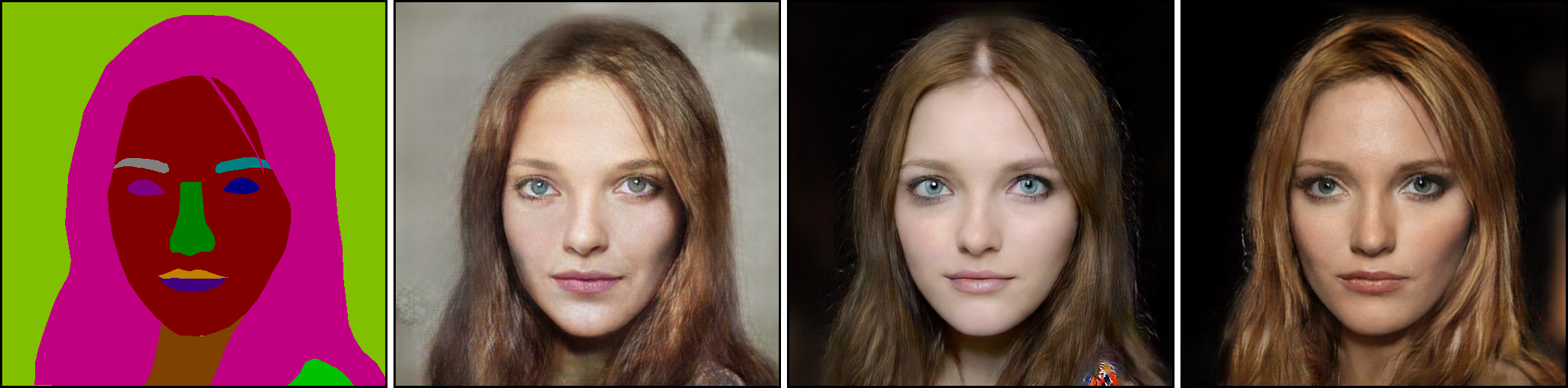} 
		\end{center}
		\vspace{-0.15in}
		\begin{tabular}{cccc}
			\hspace{0.18\columnwidth}(1) Input & \hspace{0.28\columnwidth}(2) SPADE & \hspace{0.26\columnwidth}(3) Mask-GAN & \hspace{0.24\columnwidth}(4) Ours \\
		\end{tabular}
		\caption{Visual comparison on CelebAMask-HQ.}
		\label{fig_exp2}
	\end{figure*}
	
	\begin{figure*}[!t]
		\begin{center}
			\includegraphics[width=0.95\linewidth]{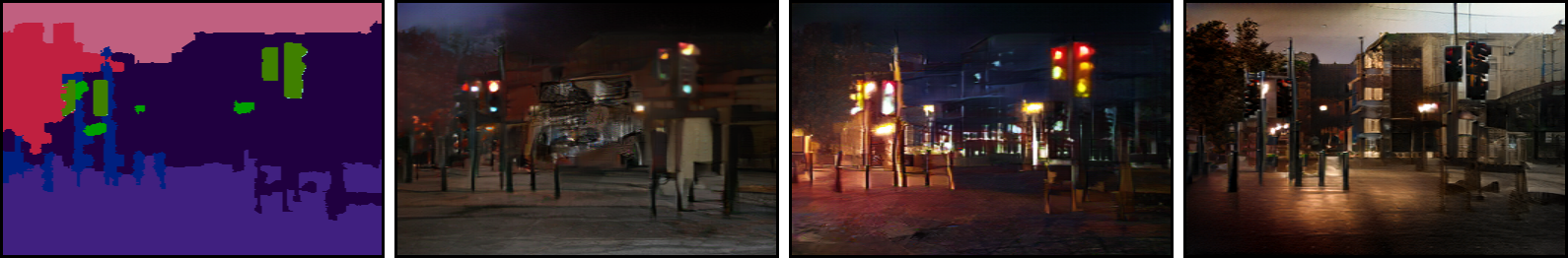}
			\includegraphics[width=0.95\linewidth]{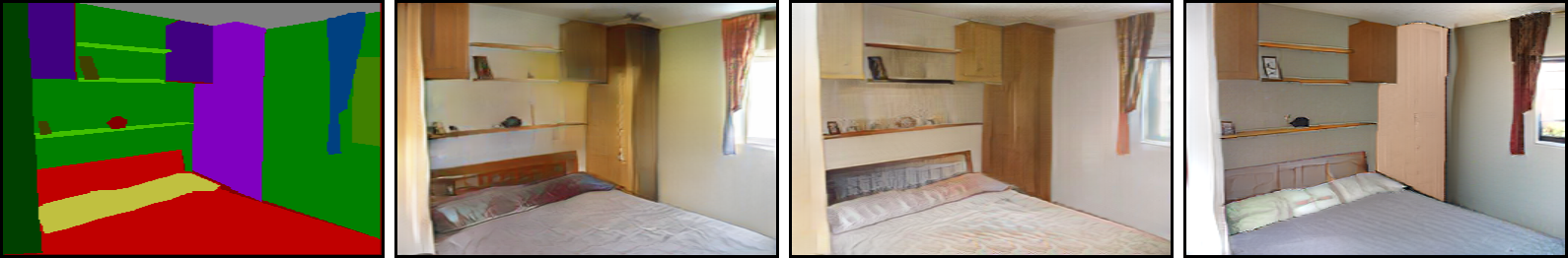} 
			\includegraphics[width=0.95\linewidth]{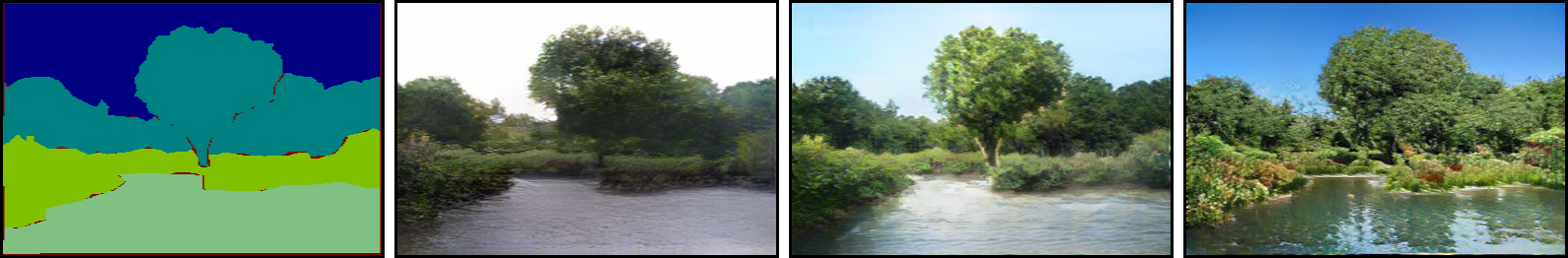} 
		\end{center}
		\vspace{-0.1in}
		\begin{tabular}{cccc}
			\hspace{0.18\columnwidth}(1) Input & \hspace{0.28\columnwidth}(2) SPADE & \hspace{0.26\columnwidth}(3) CC-FPSE & \hspace{0.26\columnwidth}(4) Ours \\
		\end{tabular}
		\caption{Visual comparison on COCO-stuff and ADE20K.}
		\label{fig_exp1}
		\vspace{-0.1in}
	\end{figure*}
	
	\subsection{Quantitative Comparison} 
	Table \ref{tb_me_evaluation1} indicates that our method yields decent performance, manifesting the effectiveness of our design of generating layout-aligned and appearance-related operators.
	On CelebAMask-HQ, our proposed SC-GAN improves face synthesis in terms of FID to 19.2, compared to FID of Spade 42.2. Our score is even lower than that computed from MaskGAN (21.4), which specializes in face manipulation utilizing \textit{ground truth} face images for style reference. In scene-related datasets, \eg\ Cityscapes, ADE20K, and COCO-Stuff, our method works nicely regarding segmentation and generation evaluation, giving non-trivial improvements compared with baselines, especially on FID.
	
	In the comparison shown in Table \ref{tb_me_evaluation1} among CRN, CC-FPSE, and our SC-GAN, SC-GAN also shows better results in terms of mIoU, Acc, and FID. It proves the necessity to learn the semantic vectors from the segmentation masks in our method, since CRN concatenates segmentation masks in every input stage and CC-FPSE parameterizes convolution by generating weights from segmentation masks.
	
	\begin{table}[t]
		\centering
		\caption{User study. Each entry gives the percentage of cases where our results are favored.}
		\label{tb_user_studies} \small
		\vspace{0.1in}
		\setlength{\tabcolsep}{1mm}{
			\begin{tabular}{c|cccc}
				\hline
				Methods & CelebAHQ & Cityscapes & ADE20K  &  COCO  \\
				\hline
				Ours $>$ SPADE & 76.00\% & 59.50\% & 66.62\% & 57.78\% \\
				Ours $>$ CC & N/A & 54.12\% & 60.28\% &  53.60\% \\
				\hline
			\end{tabular}
		}
		\vspace{-0.1in}
	\end{table}
	
	\vspace{-0.15in}
	\paragraph{User Study} Adhering to the protocol in SPADE \cite{park2019semantic}, we list our user study results in Table \ref{tb_user_studies} to compare our method with SPADE and CC-FPSE. Specifically, the subject judges which synthesizing result looks more natural corresponding to the input semantic layout. In all conditions, results from our model are more preferred by users compared with those from others, especially on CelebAMask-HQ.
	
	\subsection{Qualitative Results}
	Figures \ref{fig_exp2} and \ref{fig_exp1} give the visual comparison of our method and other baselines. 
	Our method generates natural results with less noticeable visual artifacts and more appearing details. Note that the skin of persons by our method is more photorealistic. Moreover, due to the effectiveness of semantic vectors, our method yields more consistent eye regions in Figure \ref{fig_exp2}, and even creates intriguing reflections on the water in the bottom row of Figure \ref{fig_exp1}. More visual results are given in the supplementary material.
	
	\begin{figure}[t]
		\begin{center}
			\includegraphics[width=0.95\linewidth]{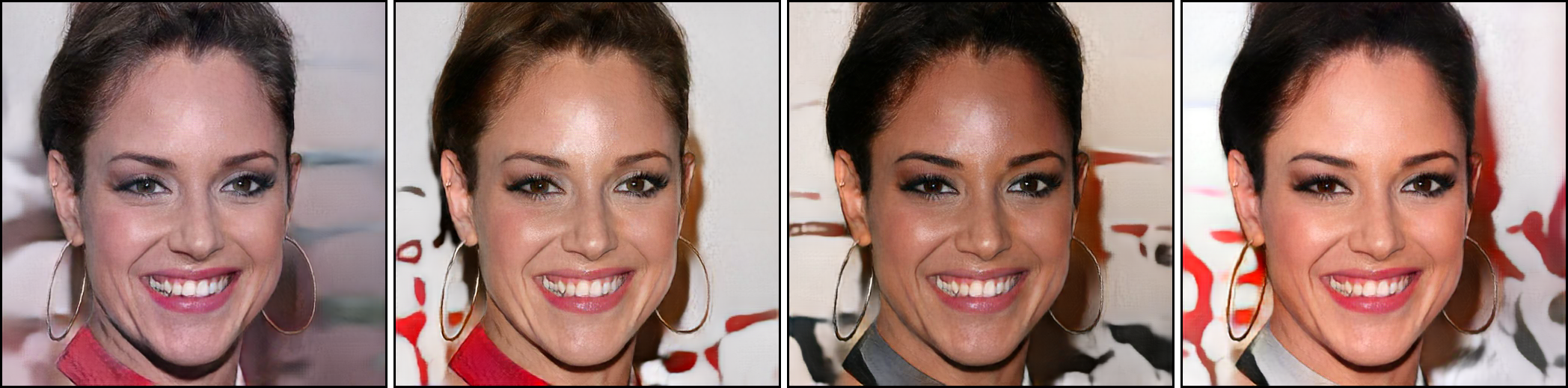} 
			\includegraphics[width=0.95\linewidth]{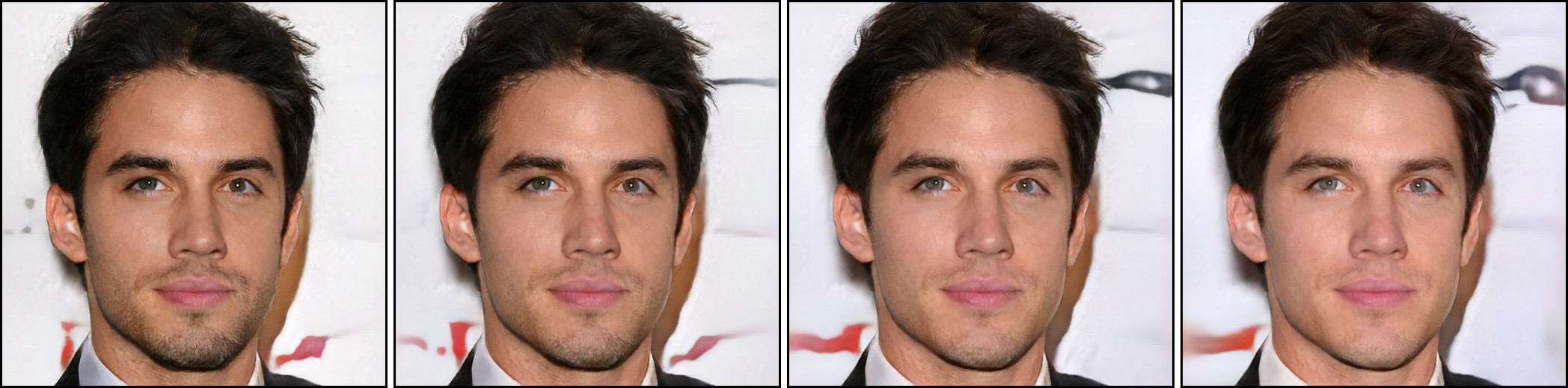}
			
		\end{center}
		\vspace{-0.1in}
		\caption{Multi-modal predictions (top row) and interpolation (bottom row) of our method on CelebAMask-HQ.}
		\label{fig_exp4}
	\end{figure}
	
	\begin{figure}[t]
		\begin{center}
			\includegraphics[width=0.95\linewidth]{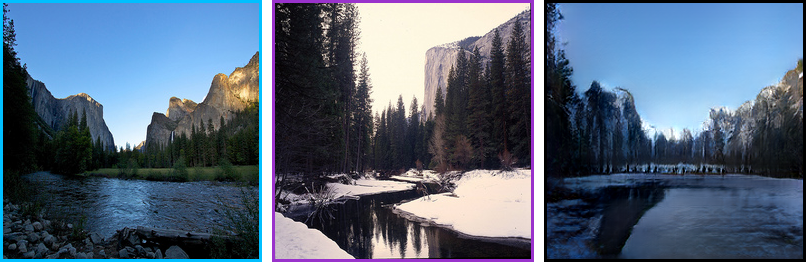}
		\end{center}
		\vspace{-0.1in}
		\caption{Unpaired image-to-image translation results from our model on summer$\rightarrow$winter. The source images, their reference ones (target images), and their corresponding translated results are marked by blue, purple, and black rectangles, respectively.}
		\label{fig_i2i}
	\end{figure}
	
	\vspace{-0.15in}	
	\paragraph{Multi-modal Outputs and Interpolation} Our method synthesizes multi-modal results from the same semantic layout using an additional encoder trained in the VAE manner. With different sampled random noise, our method gives diverse output in terms of appearance (top row in Figure \ref{fig_exp4}). Also, we can apply linear interpolation of these random vectors, achieving a smooth transition from having a beard to not with it (bottom row in Figure \ref{fig_exp4}). It validates the effectiveness of the learned manifold in our model.
	
	\begin{table}[t]
		\centering
		\caption{Impact of different operators (Op) on the generative performance on Cityscapes.}
		\label{tb_ab_ops} \small
		\vspace{0.1in}
		\begin{tabular}{c|ccccc}
			\hline
			Op & Conv+BN & Conv+SCN & SCC+BN & SCC+SCN \\
			\hline
			mIoU$\uparrow$ & 62.6 & 63.1 & 66.3  & \textbf{66.9}\\
			FID$\downarrow$ & 69.7 & 60.9 & 54.0  & \textbf{49.5}\\
			\hline
		\end{tabular}
	\end{table}
	
	\subsection{Ablation Studies} \label{sec_ab}
	We ablate the key design of our method on Cityscapes.
	
	\vspace{-0.15in}
	\paragraph{SCC vs. Standard Conv} We construct three baselines, where SRG employs conventional convolutions (Conv) and Batch normalizations (BN), Conv and SCN, SCC and BN, respectively, and SVG remains intact. For baselines using Conv, we triple its Conv number in SRG compared to SCC for fair comparisons, as $n=3$ in Eq. \eqref{eq_scc}. We use extra input (concatenating $\mathbf{V}$ to the input feature maps $\mathbf{F}$) for every Conv, to see how the usage of $\mathbf{V}$ affects the performance.
	
	From Table \ref{tb_ab_ops}, we notice SCC boosts Cityscapes synthesis from Conv in terms of mIoU and FID (mIoU: $62.6 \rightarrow 66.3$, FID: $69.7 \rightarrow 54.0$ with BN, and mIoU: $63.1 \rightarrow 66.9$, FID: $60.9 \rightarrow 49.5$ with SCN). It manifests the effectiveness of SCC. Exploiting spatially variant features ($\mathbf{V}$) by the proposed dynamic operators is more beneficial to yielded results than by static ones in this task, with a slimmer capacity (66.2M vs. 67.3M (Conv+BN)).
	
	\vspace{-0.15in}
	\paragraph{SCN vs. BN vs. SPADE} The design of SCN is also validated in Table \ref{tb_ab_ops}. SCN improves synthesis more on the generation quality (FID: $69.7 \rightarrow 60.9$ with Conv, and FID: $54.0 \rightarrow 49.5$ with SCC). Its influence on the semantic alignment (mIoU) is relatively small. Also, Conv+SCN gives better mIoU and FID compared SPADE (Table \ref{tb_me_evaluation1}), further validating the importance of SCN.
	
	\begin{table}
		\centering
		\caption{Impact of semantic vector generation loss $\mathbf{L}_s$ in Eq. \eqref{eq_svg} about the generative performance on Cityscapes.}
		\vspace{0.1in}
		\label{tb_ab_svg} \small
		\begin{tabular}{c|cccc}
			\hline
			& w/o $\mathcal{L}_s$ & w/ $\mathcal{L}_s$ ($\ell_1$) & w/ $\mathcal{L}_s$ ($\ell_2$) & w/ $\mathcal{L}_s$ (vgg) \\
			\hline
			mIoU$\uparrow$ & 60.3 & \textbf{66.9} & 63.4 & 64.2\\
			FID$\downarrow$ & 82.5 & \textbf{49.5} & 53.3  & 51.8\\
			\hline
		\end{tabular}
	\end{table}
	
	\vspace{-0.15in}
	\paragraph{Semantic Vector Generation Loss} As mentioned in Section \ref{sec_svg}, semantic vector generation loss is to improve the relationship between different semantics according to their appearance. As shown in Table \ref{tb_ab_svg}, without it in Eq. \eqref{eq_svg}, the corresponding quantitative performance degrades notably both on object alignment (mIoU: $66.9 \rightarrow 60.3$) and generation (FID: $49.5 \rightarrow 82.5$). Also, using $\ell_1$ norm in Eq. \eqref{eq_svg} is better than using $\ell_2$ norm or perceptual loss considering both alignment and generation.
	
	\begin{table}
		\centering
		\caption{Generative performance on Cityscapes regarding the number of the shared weight candidates in SCC.}
		\vspace{0.1in}
		\label{tb_ab_num}
		\begin{tabular}{c|cccc}
			\hline
			& $n=8$ & $n=4$ & $n=2$ & $n=1$ \\
			\hline
			mIoU$\uparrow$ & \textbf{67.9} & 67.3 & 64.8 & 61.2\\
			\hline
		\end{tabular}
		\vspace{-0.1in}
	\end{table}
	
	\vspace{-0.15in}
	\paragraph{Number of Shared Weight Candidates in SRG} We reduce the number of conv weight candidates in SRG while preserving that of norm candidates (fixed to 4). The corresponding segmentation results are shown in Table \ref{tb_ab_num}. With the increase of $n$, the semantic alignment becomes better. 
	
	\begin{table}
		\centering
		\caption{Different nonlinear functions affect the generative performance on Cityscapes.}
		\label{tb_ab_act} \small
		\vspace{0.05in}
		\begin{tabular}{c|ccccc}
			\hline
			Nonlinear & Sigmoid & Tanh & ReLU & None & Softmax \\
			\hline
			mIoU$\uparrow$ & 62.6  & 63.1 & 61.7 & 60.2 & \textbf{66.9}\\
			FID$\downarrow$ & 57.6 & 56.8 & 66.5 & 73.9 & \textbf{49.5}\\
			\hline
		\end{tabular}
		\vspace{-0.01in}
	\end{table}
	
	\vspace{-0.15in}
	\paragraph{Transformation of Semantic Vector Computation} We evaluate different nonlinear functions for semantic vector computation, as given in Table \ref{tb_ab_act}. Note incorporating functions on $\mathbf{V}$ is vital as mIoU and FID scores become worse without it (mIoU: 60.2, FID: 73.9), and using softmax yields the best quantitative generation performance. In our model, bounded activation functions work better than unbounded (sigmoid, tanh, softmax vs. ReLU) ones, and the normalized one performs better than the unnormalized setting (softmax vs. sigmoid and tanh).
	
	\vspace{-0.15in}
	\paragraph{With A New Discriminator and Training Tricks} We note new efforts \cite{sushko2020you} were made to enhance synthesis results by using a more effective discriminator and training approaches. Incorporating these techniques into our method could further boost our generation performance, \eg, mIoU: 69.9, FID: 47.2 on Cityscapes, and mIoU: 49.1, FID: 27.6 on ADE20K, as given in Section 2.1 of the supp. material.
	
	\begin{table}
		\centering
		\caption{Unpaired image-to-image translation evaluation on summer-to-winter dataset.}
		\label{tb_style} 
		\vspace{0.05in}
		\begin{tabular}{c|ccc}
			\hline
			Methods & MUNIT \cite{huang2018munit} & DMIT \cite{yu2019multi} & Ours\\
			\hline
			FID$\downarrow$ & 118.225 & 87.969 & \textbf{82.882} \\
			IS$\uparrow$ & 2.537 & 2.884 &  \textbf{3.183}\\
			\hline
		\end{tabular}
		\vspace{-0.1in}
	\end{table}
	
	\subsection{Unpaired Image-to-Image Translation}  
	Due to our semantic encoding and stylization design, our model can also be applied to unpaired image-to-image (i2i) translation with minor modification. This modified framework is given in the supp. material. 
	Table \ref{tb_style} shows the quantitative evaluation about our model along with unpaired i2i baselines MUNIT \cite{huang2018munit} and DMIT \cite{yu2019multi} on Yosemite summer-to-winter dataset. The superiority of FID and IS from our model demonstrates its generality. Visual results are given in Figure \ref{fig_i2i}.
	
	\section{Concluding Remarks}
	In this paper, we have presented a new method to represent visual content in a semantic encoding and stylization manner for generative image synthesis. Our method introduces appearance similarity to semantic-aware operations, proposing a novel spatially conditional processing for both convolution and normalization. It outperforms the compared popular synthesis baselines on several benchmark datasets both qualitatively and quantitatively. 
	
	This new representation is also beneficial to unpaired image-to-image translation. We will study its applicability to other generation tasks in the future.
	
	\vspace{-0.1in}
	
	{\small
		\bibliographystyle{ieee_fullname}
		\bibliography{egbib}
	}

\newpage

\begin{appendices}
	\section*{Appendix}
	
	In this supplementary file, our descriptions contain the following components:
	\begin{itemize}
		\item Detailed configuration of our proposed SC-GAN and the implementation of the given spatially conditional convolution and normalization.
		\item The synthesis performance of our generator trained with a more effective discriminator and other tricks.
		\item The specification of how to apply our framework to unpaired image-to-image translation and the corresponding visual results.
		\item The limitations and some failure cases of our method.
	\end{itemize}
	
	\section{Network Architectures}
	SC-GAN consists of Semantic Vector Generator (SVG) and Semantic Render Generator (SRG). Their detailed designs are given blew. For convenience, we suppose Conv(k, s, c) indicates a convolutional operation whose kernel size, stride size, and output channel number of the used convolution are k, s, and c, respectively. The dilation ratio and padding size of Conv(k, s, c) are both set to 1. $\uparrow$ and $\otimes$ denotes a $2 \times$ bilinear upsampling and concatenation (along with the channel dimension) operations, respectively. Besides, SCResBlock(k, s, c) denotes a residual block variant using spatially conditional convolution (SCC) and normalization (SCN). Its schematic illustration is given in Figure 4 of our paper. We use $\mathbf{y}[+\mathbf{x}]$ to indicate an extra input $\mathbf{x}$ of the current operation $\mathbf{y}$. For example, 
	SCResBlock(3,1,512)$[+\mathbf{V}_i]$ indicates the semantic vectors $\mathbf{V}_i$ (in feature maps form) is also incorporated into SCResBlock(3,1,512) for generating dynamic operators.
	
	\paragraph{SVG}: $\mathbf{S}\downarrow$ $\rightarrow$ Conv(3,1,512) $\rightarrow$ LReLU $\rightarrow$ Conv$_1$(3,1,512)  $\rightarrow \uparrow$ $\rightarrow$ $\otimes (\mathbf{S}\downarrow)$ $\rightarrow$ LReLU $\rightarrow$ Conv (3,1,256) $\rightarrow$ LReLU $\rightarrow$ Conv$_2$(3,1,256) $\rightarrow \uparrow$ $\rightarrow$ $\otimes (\mathbf{S}\downarrow)$ $\rightarrow$ LReLU $\rightarrow$ Conv(3,1,128) $\rightarrow$ LReLU $\rightarrow$ Conv$_3$(3,1,128) $\rightarrow \uparrow$ $\rightarrow$ $\otimes (\mathbf{S}\downarrow)$ $\rightarrow$ LReLU $\rightarrow$ Conv(3,1,64) $\rightarrow$ LReLU $\rightarrow$ Conv$_4$(3,1,64) $\rightarrow \uparrow$ $\rightarrow$ $\otimes (\mathbf{S}\downarrow)$ $\rightarrow$ LReLU $\rightarrow$ Conv(3,1,32) $\rightarrow$ LReLU $\rightarrow$ Conv$_5$(3,1,32) $\rightarrow \uparrow$ $\rightarrow$ $\otimes (\mathbf{S}\downarrow)$ $\rightarrow$ LReLU $\rightarrow$ Conv(3,1,32) $\rightarrow$ LReLU $\rightarrow$ Conv$_6$(3,1,32) $\rightarrow \uparrow$ $\rightarrow$ $\otimes (\mathbf{S})$ $\rightarrow$ LReLU $\rightarrow$ Conv(3,1,16) $\rightarrow$ LReLU $\rightarrow$ Conv(3,1,3) $\rightarrow$  Hardtanh $\rightarrow$ $f_\text{V}^\text{out}(\mathbf{S})$,\\
	where $\mathbf{S}$ indicates the input segmentation mask.
	
	\paragraph{SRG}: $z$ $\rightarrow$ SCResBlock(3,1,512) [+$\mathbf{V}_1$] $\rightarrow \uparrow$ $\rightarrow$ SCResBlock(3,1,512) [+$\mathbf{V}_2$] $\rightarrow \uparrow$ $\rightarrow$ SCResBlock(3,1,512) [+$\mathbf{V}_2$] $\rightarrow \uparrow$ $\rightarrow$ SCResBlock(3,1,256) [+$\mathbf{V}_3$] $\rightarrow \uparrow$ $\rightarrow$ SCResBlock(3,1,128) [+$\mathbf{V}_4$] $\rightarrow \uparrow$ $\rightarrow$ SCResBlock(3,1,64) [+$\mathbf{V}_5$] $\rightarrow \uparrow$ $\rightarrow$ SCResBlock(3,1,32) [+$\mathbf{V}_6$] $\rightarrow \uparrow$ $\rightarrow$ Conv (k3c3) $\rightarrow$  Hardtanh $\rightarrow$ $\mathbf{\hat{I}}$,\\
	where $z$ is sampled from a standard normal distribution, $\mathbf{V}_i$ $\dashleftarrow$ $\text{Conv}_i$ from SVG, and $\dashleftarrow$ denotes the adaptive pooling operation (in channel dimension).

	\begin{figure*}[!h]
		\begin{center}
			\centering
			\includegraphics[width=0.9\linewidth]{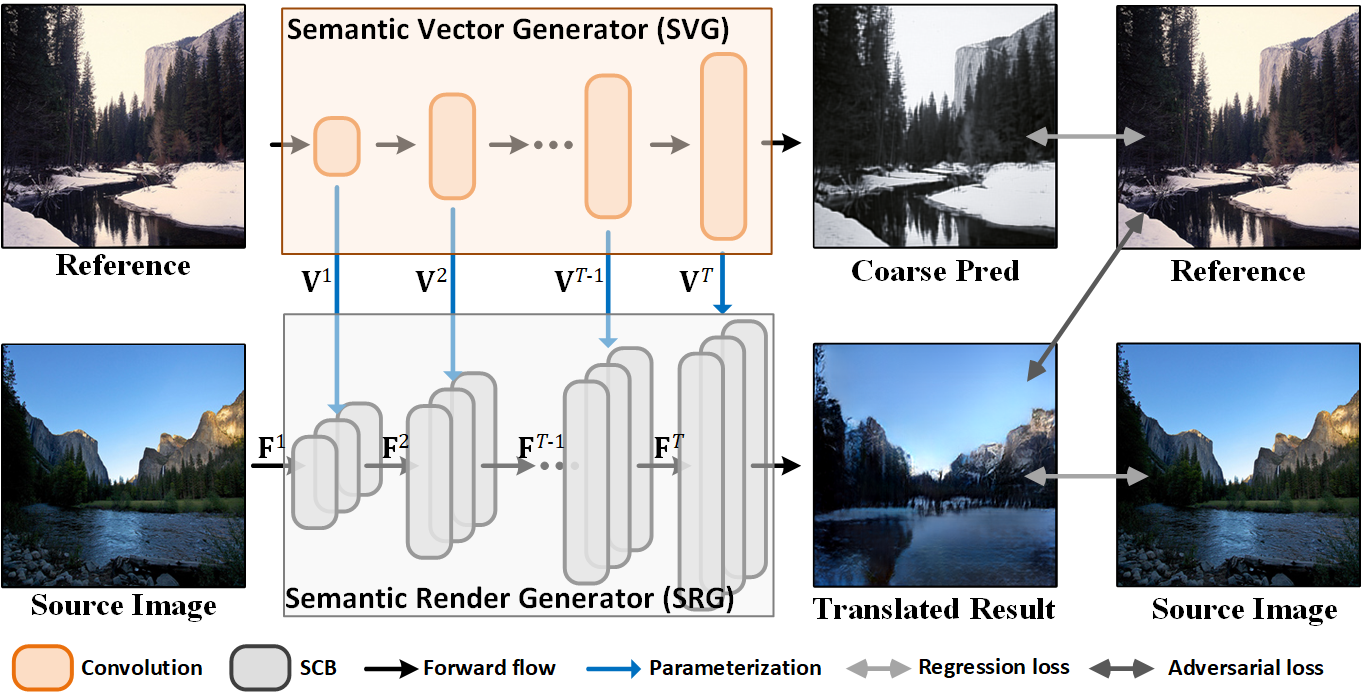} 
		\end{center}
		\vspace{-0.1in}
		\caption{The framework of SC-GAN for unpaired image-to-image translation.}
		\label{fig_frame}
	\end{figure*}
	\subsection{Implementation of Key Components}
	
	The pseudo codes to realize the proposed spatially conditional convolution (SCC) and normalization (SCN) are given in Alg \ref{alg1} and \ref{alg2}, respectively. In general, they are designed that the regional parameterized weights are generated by combining a group of candidates, and the manner how they are combined is indicated by the fed semantic vectors (e.g. the following $\mathbf{V}$).
	
	\begin{algorithm}
		\caption{The pseudo code of SCC (PyTorch style)} \label{alg1}
		\begin{algorithmic}[1]
			\renewcommand{\algorithmicrequire}{\textbf{Input:}}
			\renewcommand{\algorithmicensure}{\textbf{Output:}}
			\Require Input feature maps $\mathbf{F} \in \mathcal{R}^{b \times c_\text{in} \times h \times w}$ and the generated semantic vectors $\mathbf{V} \in \mathcal{R}^{b \times n \times h \times w}$,  and learnable parameter candidates $\{\mathbf{k}_i\}$ where $\mathbf{k}_i \in \mathcal{R}^{c_\text{in} \times ks \times ks \times c_\text{out}}$, and $c_\text{in}$, $c_\text{out}$, and $ks$ denote the input, output channel number, and kernel size.
			\Ensure The convolved feature maps $\mathbf{\hat{F}} \in \mathcal{R}^{b \times c_\text{out} \times h \times w}$.
			\State $\mathbf{V} = \mathbf{V}.\text{unsqueeze}(2)$ \textcolor{gray}{\# shape: $b \times n \times 1 \times h \times w$}
			\State $\text{out} = [0] * n$
			\For{$i=1$ to $n$}
			\State $\text{out}[i] = \mathbf{k}_i(\mathbf{F}).\text{unsqueeze}(1)$
			\EndFor
			\State $\mathbf{\hat{F}} = \text{torch.cat}(\mathbf{F}, \text{dim}=1)$ \textcolor{gray}{\# shape: $b \times n \times c_\text{out} \times h \times w$}
			\State $\mathbf{\hat{F}} = \mathbf{\hat{F}} * \mathbf{V}$
			\State $\mathbf{\hat{F}} = \text{torch.sum}(\mathbf{\hat{F}}, \text{dim}=1)$ \textcolor{gray}{\# shape: $b \times c_\text{out} \times h \times w$}
		\end{algorithmic}
	\end{algorithm}
	
	\begin{algorithm}
		\caption{The pseudo code of SCN (PyTorch style)} \label{alg2}
		\begin{algorithmic}[1]
			\renewcommand{\algorithmicrequire}{\textbf{Input:}}
			\renewcommand{\algorithmicensure}{\textbf{Output:}}
			\Require Input feature maps $\mathbf{F} \in \mathcal{R}^{b \times c_\text{in} \times h \times w}$ and the generated semantic vectors $\mathbf{V} \in \mathcal{R}^{b \times n \times h \times w}$,  and learnable parameter candidates $\mathbf{A}$, where $\mathbf{A} \in \mathcal{R}^{n \times 2c_\text{in}}$.
			\Ensure The normalized feature maps$\mathbf{\hat{F}} \in \mathcal{R}^{b \times c_\text{in} \times h \times w}$.
			\State $\mathbf{V} = \mathbf{V}.\text{permute}(0,2,3,1).\text{contiguous()}$
			$.\text{view}(-1, n)$ \textcolor{gray}{\# shape: $b  \times n \times h \times w \rightarrow bhw \times n$}
			\State $\mathbf{\hat{F}} = \text{BN}(\mathbf{F})$ \textcolor{gray}{\# shape: $b \times c_\text{in} \times h \times w$}
			\State $\mathbf{A} = \text{torch.matmul}(\mathbf{V}, \mathbf{A}).\text{view}(b, h, w, -1).\text{permute}$
			$(0, 3, 1, 2).\text{contiguous}()$ \textcolor{gray}{\# shape: $b  \times 2c_\text{in} \times h \times w$}
			\State $\mathbf{m}, \mathbf{s}=\text{torch.split}(\mathbf{A}, n, \text{dim}=1)$
			\State $\mathbf{\hat{F}} = \mathbf{\hat{F}} * (1+\mathbf{s}) + \mathbf{m}$
		\end{algorithmic}
	\end{algorithm}
	
	\section{More Experimental Results and Analysis}
	
	
	\begin{table}[t]
		\centering
		\caption{Quantitative results on the validation sets of Cityscapes and ADE20K from different methods. Ours w OASIS denotes our generator is trained with the discriminator and other techniques from OASIS \cite{sushko2020you}.}
		\label{tb_eval}
		\vspace{0.1in}
		\setlength{\tabcolsep}{1mm}{
			\begin{tabular}{c|cccc}
				\toprule
				\multirow{2}*{Method} &  \multicolumn{2}{c}{Cityscapes} & \multicolumn{2}{c}{ADE20K}\\
				~  & mIoU $\uparrow$ & FID $\downarrow$ & mIoU $\uparrow$ & FID $\downarrow$ \\
				\midrule
				
				SPADE \cite{park2019semantic}  & 62.3  & 71.8 & 38.5  & 33.9 \\
				CC-FPSE \cite{liu2019learning}  & 65.6  & 54.3 & 43.7 & 31.7 \\
				OASIS \cite{sushko2020you} & 69.3 & 47.7 & 48.8 & 28.3\\
				Ours & 66.9 & 49.5 & 45.2 & 29.3\\
				Ours w OASIS  & \textbf{69.9} & \textbf{47.2} & \textbf{49.1} & \textbf{27.6}\\
				\bottomrule
			\end{tabular}
		}
		\vspace{-0.1in}
	\end{table}
	
	\subsection{Performance with a Stronger Discriminator and Training Tricks}
	We evaluate the compatibility between our proposed generator and a newly introduced discriminator \cite{sushko2020you}, along with some effective training techniques. Sushko \etal\ presented a powerful discriminator exploiting semantic layouts (OASIS) by semantic segmentation loss. Their approach further strengthens GAN training by 1) balancing the class weights by their frequencies, 2) removing VGG loss, and 3) exponential moving average (EMA) model merging (for generator). During training, they mask generated regions with a random class and add 3D random noise to all input segmentation masks to enhance local detail synthesis. By integrating their techniques (except 3D random noises), our model can be further improved over performance, as given in Table \ref{tb_eval}. Compared with OASIS \cite{sushko2020you}, our proposed generator yields better quantitative results with a smaller capacity (66.2M (ours) vs. 94M (OASIS)).
	
	\subsection{Unpaired Image-to-image Translation}
	As we claimed in the paper, our method is also applicable to unpaired image-to-image translation applications with minor modifications. Its corresponding framework is presented in Figure \ref{fig_frame}. Specifically, changing the input of SRG from the random noise to the downsampled image from the source domain, then the output of SRG should be close to its input one in semantic layout (using perceptual loss), and similar to images from the target domain in style (texture, detail, etc). The input of SVG is set to the image from the target domain, and it still regresses to the input image like an autoencoder. Figure \ref{fig_i2i} gives visual results from our model on summer$\rightarrow$winter dataset \cite{zhu2017unpaired}, in which our proposed method can alter the source image style by changing its color and texture according to the reference.

	\subsection{Limitations and Failure Cases}
	Although our design enhances generation performance by explicitly learning the relations between different semantics, it may fail to fully recover the intrinsic geometry in the original image when the given segmentation map is short of such cues (Figure \ref{fig_fail}). This is a common issue in numerous generative models, usually addressed by introducing extra modal (\eg\ depth) or multi-view data. 
	
	Moreover, the explicit modeling between semantics by their appearances may lead to creating undesired objects/stuff in the target semantic region, as given in Figure \ref{fig_fail}. Though the details in building regions are vivid, unexpected plants are synthesized in these areas. We suppose it is caused by that our model learns such symbiotic bias between these two kinds of stuff in the training data. Utilizing segmentation masks to further constrain semantic vectors (\eg\ adding semantic segmentation loss to SVG) may address this issue. We will study it in the future.

	\begin{figure*}[!t]
		\begin{center}
			\centering
			\setlength{\tabcolsep}{1mm}{
				\begin{tabular}{cc}
					\includegraphics[width=0.48\linewidth]{experiments/summer2winter_extracted/0.png} &
					\includegraphics[width=0.48\linewidth]{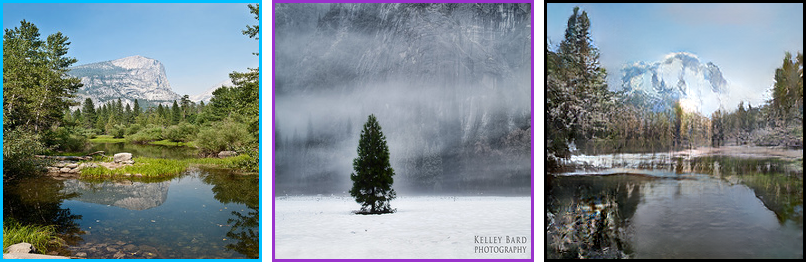} \\
					\includegraphics[width=0.48\linewidth]{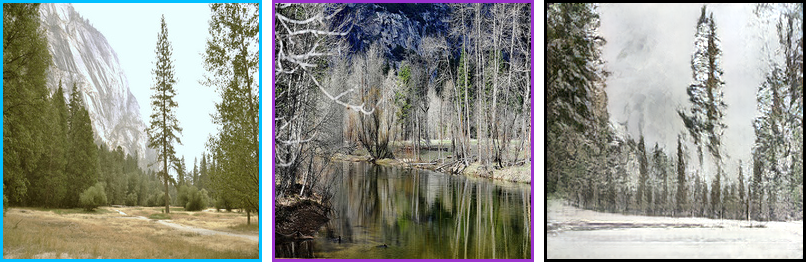} &
					\includegraphics[width=0.48\linewidth]{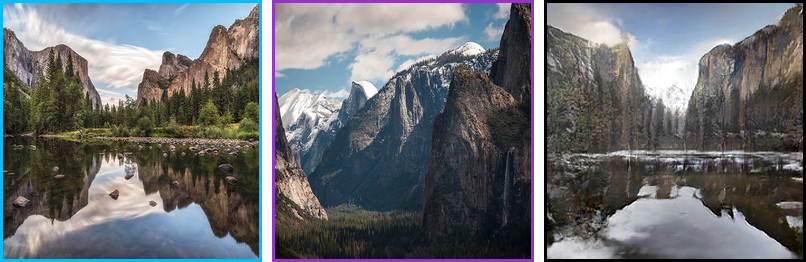} \\
				\end{tabular}
			}
		\end{center}
		\vspace{-0.1in}
		\caption{Unpaired image-to-image translation results from our model on summer$\rightarrow$winter. The source images, their reference ones (target images), and their corresponding translated results are marked by blue, purple, and black rectangles, respectively.}
		\label{fig_i2i}
	\end{figure*}
	
	\begin{figure*}[t]
		\begin{center}
			\includegraphics[width=0.95\linewidth]{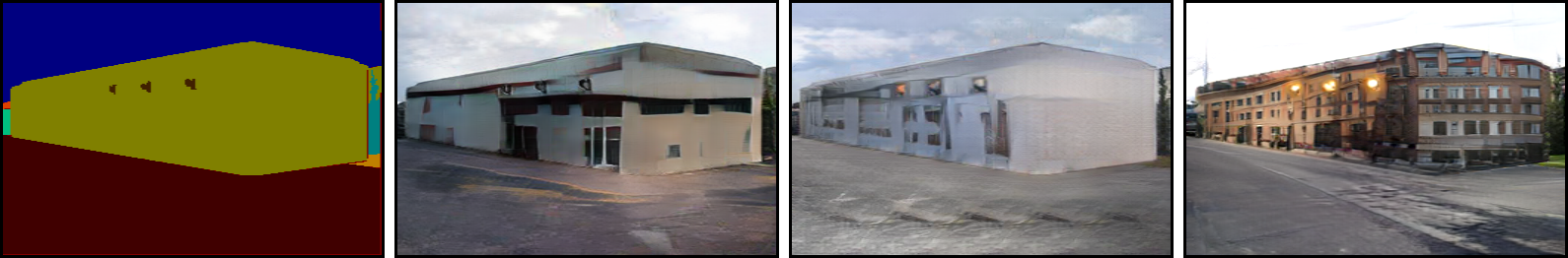} 
			\vspace{0.02in} 
			\includegraphics[width=0.95\linewidth]{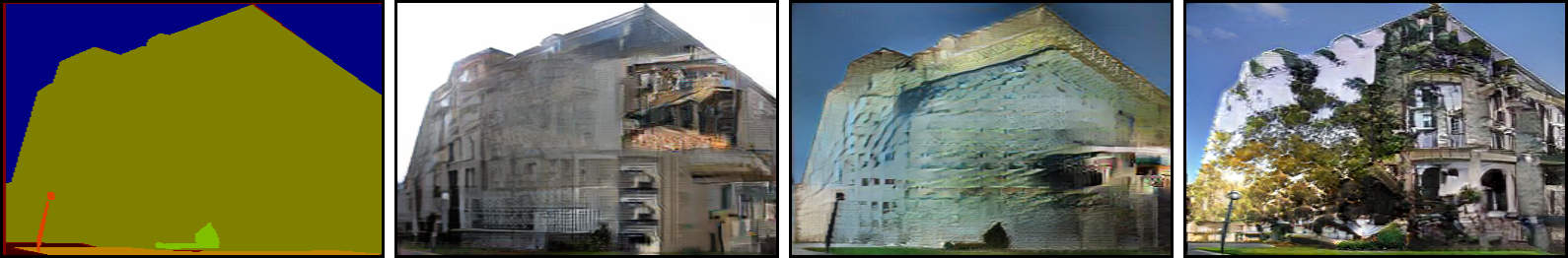} 
			
		\end{center}
		\vspace{-0.1in}
		\begin{tabular}{cccc}
			\hspace{0.15\columnwidth}(1) The input & \hspace{0.25\columnwidth}(2) SPADE & \hspace{0.25\columnwidth}(3) CC-FPSE & \hspace{0.25\columnwidth}(4) Ours. \\
		\end{tabular}
		\caption{Failure cases. The top row: failing to add dimension and depth in facade. The bottom row: introducing undesired objects in the given semantic regions.}
		\label{fig_fail}
	\end{figure*}
\end{appendices}

\end{document}